# Efficient AI Model Deployment Using Quantization Analysis Tool

Dwith Chenna
Electrical and Computer Engineering Department
University of Maryland
College Park, MD, USA
cyndwith@ieee.org

Kanishka Macherla
Electrical and Computer Engineering Department
University of Maryland
College Park, MD, USA
*kanishkamacherla@gmail.com*

***Abstract*—As deep learning models are increasingly deployed on resource-constrained devices, the demand for efficient model optimization techniques continues to grow. Effective deployment of AI models on edge and low-power platforms requires optimization methods that reduce model size and computational cost while maintaining high accuracy. This paper presents *Quantization Analysis Tool*, a practical system designed to streamline quantization workflows and support performance-efficient model deployment. Built on the ONNX framework for broad interoperability, the tool provides detailed layer-wise sensitivity analysis, visualization of weight and activation distributions, and insights to guide precision selection. By identifying layers that are resilient or sensitive to reduced precision, the tool enables developers to make informed trade-offs between model size, latency, and accuracy. Experimental evaluations across multiple neural network architectures demonstrate that the tool effectively improves the quantized accuracy, leading to improved efficiency in real-world deployment scenarios. The tool also provides developers valuable insights into the effects on quantization on the model and its accuracy. This work highlights the tool's capabilities, practical applications, and its role in enabling efficient AI model deployment through robust quantization analysis.**

***Keywords*—*Model Optimization, Model Quantization, ONNX, Quantization Analysis Tool***

## I. Introduction

Artificial Intelligence (AI) has been advancing at a very rapid pace, leading the technology revolution across different fields. Deep Neural Networks have found applications in a wide variety of applications from image recognition to generative AI applications, offering a versatile solution for many real-world complex problems. These models are growing larger and increasing in computational intensity, leading to challenges for deployment on hardware with limited resources. Deploying deep neural networks (DNNs) on resource-constrained platforms—such as embedded systems, mobile devices, and edge accelerators—presents significant challenges due to limited computational capacity, memory bandwidth, and power budgets. While state-of-the-art models continue to grow in size and complexity, practical deployment environments increasingly demand lightweight, latency-efficient, and energy-aware solutions.

Quantization [1,2], has emerged as a popular technique to reduce the model size for efficient inferences with lower memory footprint and power consumption. The quantization process involves reducing the precision of model parameters (i.e. weights and activations) to decrease the computational complexity. The quantization process adds quantization error into the model, which can lead to significant reduction in model accuracy. It is crucial to understand the nature and distribution of these quantization errors within the AI model and its impact on the overall performance of the model. Analyzing, tuning and profiling these quantization trade-offs to find the optimal quantization scheme for model accuracy, model size and performance remains a challenging and complex task. To address these challenges, this paper presents the Quantization Analysis Tool, a systematic framework for evaluating quantization effects across individual layers of a model. Built on top of the ONNX (Open Neural Network Exchange) framework [3], the tool provides a unified platform for model analysis, enabling consistent quantization analysis regardless of the model's originating framework. The proposed tool performs fine-grained sensitivity analysis, generates detailed visualizations of weight and activation distributions, and supports developers in making informed decisions regarding precision assignment. Through these capabilities, the tool aims to enhance the reliability and efficiency of quantized model deployment on edge devices. Enabling the user to profile the quantized model, before deployment to identify problematic operations and select the optimal quantization scheme of quantized model deployment with improved accuracy and performance.

In this paper, we propose a Model Quantization Analysis Tool for evaluating quantized ONNX models [4], particularly useful for deploying models on resource-constrained edge devices such as CPUs, GPUs, and NPUs. Section II provides background and related work on model quantization and introduces the ONNX framework. Section III discusses the methodology and implementation details, including the ONNX model parser, quantization configuration options, and visualization tools for layer-wise error analysis and weight/activation distributions. Sections IV and V present the experimental results and offer a detailed discussion of insights derived from experiments on popular models such as ResNet, MobileNet, and EfficientNet.

## II. Background

Quantization refers to the process of reducing the numerical values of model parameters like weights/activations from high-precision to lower-precision i.e. from 32/16-bit float point representation to 8/4-bit integer representation. This technique is used to reduce the model size of already trained models reducing the compute and memory requirements, thereby allowing faster inference on AI accelerators and reducing power consumption [5]. However, the process of quantization often introduces quantization error into the AI models, which can significantly impact the model prediction accuracy [6]. While accuracy loss is expected due to reduction in quantization error from reduction in precision, this reduction needs to be within the practical limits for deployment.

When the model accuracy decreases significantly after quantization, it is possible to reduce the quantization error by using high bit-width to increase the precision. Using higher precision like 16/32 bits for the complete model can significantly increase the compute and memory requirements. Many advanced quantization techniques use adaptive bit-width [7] or selective quantization [8,9] to provide the best trade-off between accuracy and performance. Sometimes, the loss in accuracy can be attributed to few problematic layers which have significantly higher quantization error or limited quantization support. To properly analyze the models for such behaviors, allowing users to better understand the optimal quantization strategy for model quantization, we introduce the Quantization Analyzer Tool, a framework that allows users to identify the problematic nodes/operations in the model by analyzing both the float and fully quantized models. The tools present a stepwise approach, where the quantized model is analyzed for layer-wise / operation-wise quantization error to determine the best approach to quantize the model. It generates a visual report of the summarization quantization analysis, making it intuitive for the user to understand the challenges for specific model quantization.

### A. *Model Quantization*

Quantization in deep neural networks (DNNs) refers to the process of lowering the numerical precision of model parameters—such as weights, activations, and biases—from formats like 32-bit floating point to more compact representations such as 8-bit integers. Doing so reduces the model's memory footprint, speeds up inference, and can significantly improve energy efficiency. These benefits, however, often come with a drop in model accuracy. Although a small reduction in accuracy is usually acceptable, the decline must stay within limits that still make the model usable in real applications. Typical quantization approaches include i) *Post-Training Quantization (PTQ)* [2,6] which Simplifies deployment but may degrade accuracy on sensitive layers. ii) *Quantization-Aware Training (QAT)* [6] provides higher accuracy but is computationally expensive and harder to integrate into existing workflows. A key issue is that different layers exhibit varying sensitivity to quantization. Uniform quantization strategies often result in sub-optimal performance, making layer-wise analysis crucial.

When the loss in accuracy becomes too large, practitioners frequently turn to selective quantization [9]. In this approach, only certain layers are quantized, or different layers are assigned different bit widths. While this can help preserve accuracy, it is generally ad-hoc and lacks a systematic method for balancing accuracy against computational efficiency. As a result, determining the best quantization strategy becomes challenging. Selective quantization is also particularly valuable when the target hardware cannot support operations at specific lower precisions, making full-model quantization infeasible. To facilitate a systematic analysis, we build the Quantization Analysis Tool using ONNX framework, which serves is widely adapted for AI model deployment.

### B. *ONNX Framework*

The Open Neural Network Exchange (ONNX) [3] framework is an open standard that enables machine learning models to be transferred seamlessly across different training frameworks and deployment runtimes. By representing models as computation graphs with standardized operators, ONNX ensures consistency and portability, allowing models trained in systems like PyTorch or TensorFlow to run efficiently on a wide range of hardware and software platforms.

This interoperability makes ONNX an excellent foundation for a quantization analysis tool. Its graph-based model format, explicit operator definitions, and built-in quantization operators (e.g., Quantize, Dequantize) allow easy inspection of layer precision and facilitate experimentation with different quantization schemes. Combined with ONNX Runtime's quantization APIs and hardware-accelerated backends, developers can simulate and evaluate quantization effects in a flexible, framework-agnostic manner.

## III. Methodology

The user interface for the *Quantization Analysis Tool* is shown in Fig. 1. The analysis tool is designed as a modular system that integrates user interaction, ONNX-based model parsing, statistical quantization analysis, and visual reporting within a unified workflow. The methodology described below outlines the full operational pipeline, from model ingestion through detailed quantization assessment and generation of quantitative insights.

The ONNX Quantization Analysis Tool consists of the following modules:

### A. *User Interface*

The tool provides a browser-based interface that enables users to upload any ONNX-compliant deep learning model. Upon loading, the model is parsed to verify structural validity, identifies list operators, and prepares the computational graph for analysis. A summary of the model—operator types, parameter counts, tensor shapes, and estimated FLOPs—is automatically generated to give users an overview of the computational and structural characteristics prior to quantization. The interface also allows users to initiate the quantization process by selecting the required configuration, after which the quantization analysis tools orchestrate all backend computations.

### B. *ONNX Model Parsing*

Once the model is uploaded, the backend uses the ONNX Graph APIs to perform a structured analysis of the computational graph. This stage involves enumerating all graph nodes, operator attributes, and dataflow dependencies, as well as extracting model weights, biases, and other learned parameters. Activation shapes are inferred using ONNX shape-inference utilities, and per-layer as well as total parameter counts are computed to support detailed model profiling.

In addition, the system estimates FLOPs using operator-specific formulas for convolution, matrix multiplication, attention mechanisms, and other computationally intensive layers. This profiling phase produces the foundational metadata required to understand the model architecture before quantization calibration and sensitivity evaluation. The extracted structural information also drives the generation of the model summary presented to the user as shown in Fig 1.

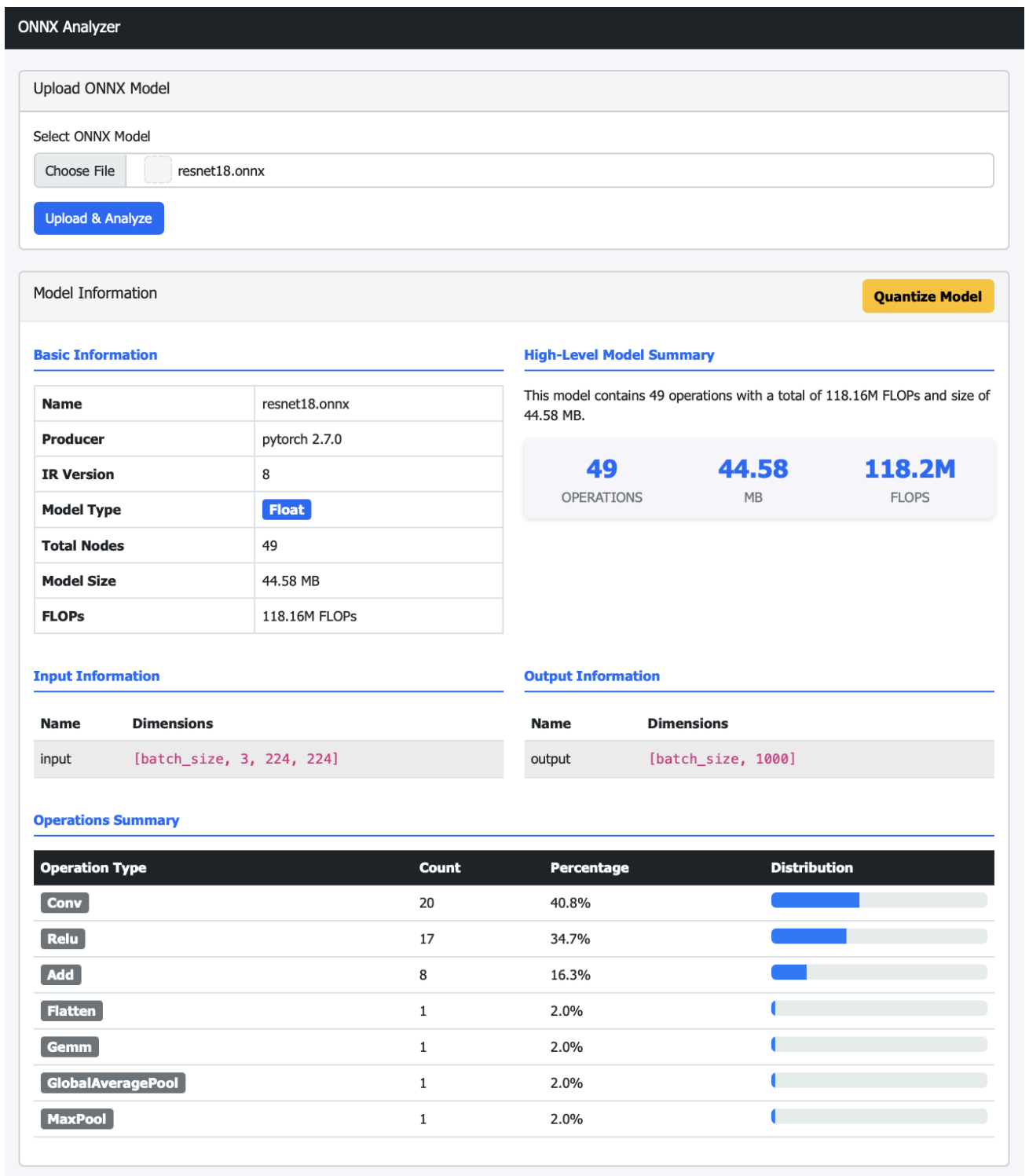

Fig. 1. Showing the UI interface with model summary information for ResNet model

## C. Model Quantization Configuration

Quantization begins with a calibration phase, which determines appropriate quantization scales and zero-points for activations and weights. The tool uses a curated subset of the ImageNet dataset, selected for its diverse feature distribution and widespread use in industry-standard quantization workflows. During calibration, the FP32 model is executed on representative batches drawn from the ImageNet calibration dataset to capture realistic activation behavior. As the model processes these inputs, activation statistics—such as minima, maxima, and optional clipping thresholds which are decided by the calibration method—are recorded for each tensor to characterize the dynamic range of values encountered during inference [6]. These collected statistics are subsequently converted into quantization parameters, including scales and zero-points, which may be computed either per-tensor or per-channel depending on the selected quantization configuration. This calibration procedure ensures that the quantized model accurately reflects the numerical distributions present in real data, thereby reducing saturation and improving overall precision. The calibration pipeline ensures that quantization ranges are aligned with the natural distribution of weights and activations, thereby reducing quantization-induced numerical saturation. The quantization configuration allows the user to select the data types (8/16 bit), calibration dataset, validation dataset and calibration method as shown in Fig 2. After selecting the appropriate quantization configuration, the tool generates the quantization analysis results for visualization and reports in the user interface.

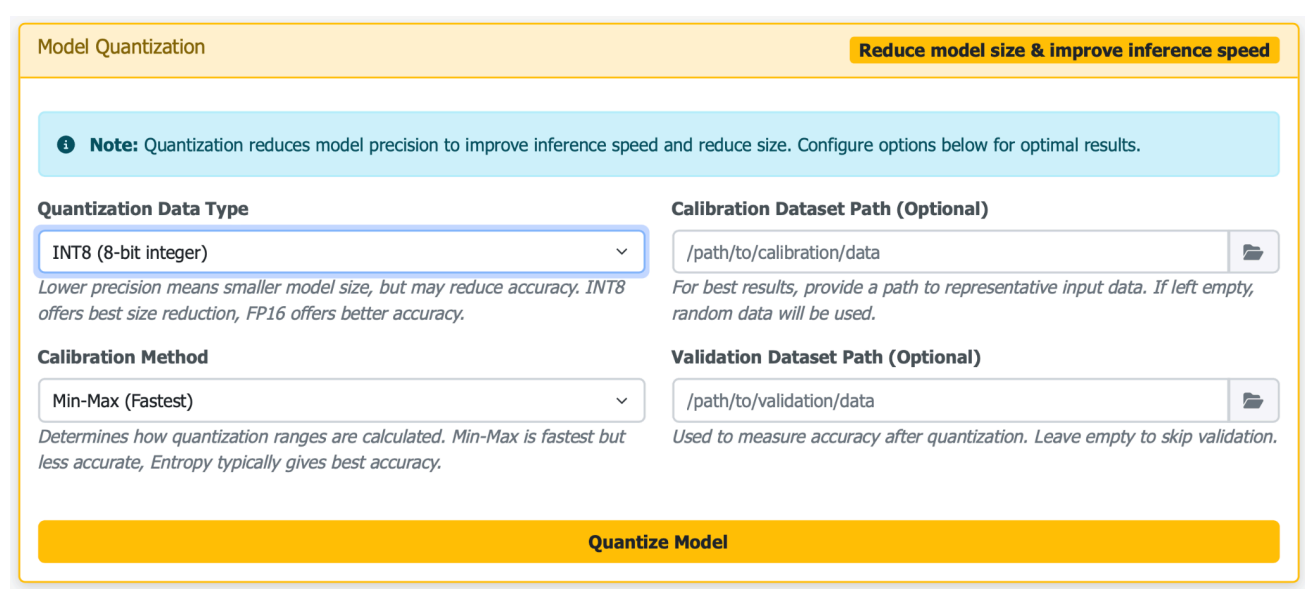

Fig. 2. Model quantization configuration with data type, calibration dataset, calibration method and validation dataset

## D. Layer-wise sensitivity analysis

Layer-wise sensitivity analysis is a core capability of the tool and enables principled identification of precision-critical operations/layers in the model. In this approach, baseline FP32 inference is first executed to establish a reference output signature. Then, each layer in the model is individually quantized while the rest of the network remains in FP32, allowing the tool to isolate the contribution of that specific layer to overall accuracy. The partially quantized model is evaluated on the calibration dataset, and the resulting accuracy drop or output deviation is measured relative to the FP32 baseline. This process provides a clear, quantitative view of which layers are most sensitive to reduced precision and therefore require special handling during deployment.

This produces a sensitivity profile that characterizes each layer's impact on end-to-end model fidelity. Layers with high sensitivity values are flagged as unsuitable for aggressive quantization, whereas robust layers with minimal degradation are recommended for lower-precision configurations. The resulting rankings guide developers in selectively applying mixed-precision quantization strategies. The tool also provides the sensitivity analysis of different operations in the topological order from top to bottom, with each operation classified as low, medium or high sensitive to quantization.

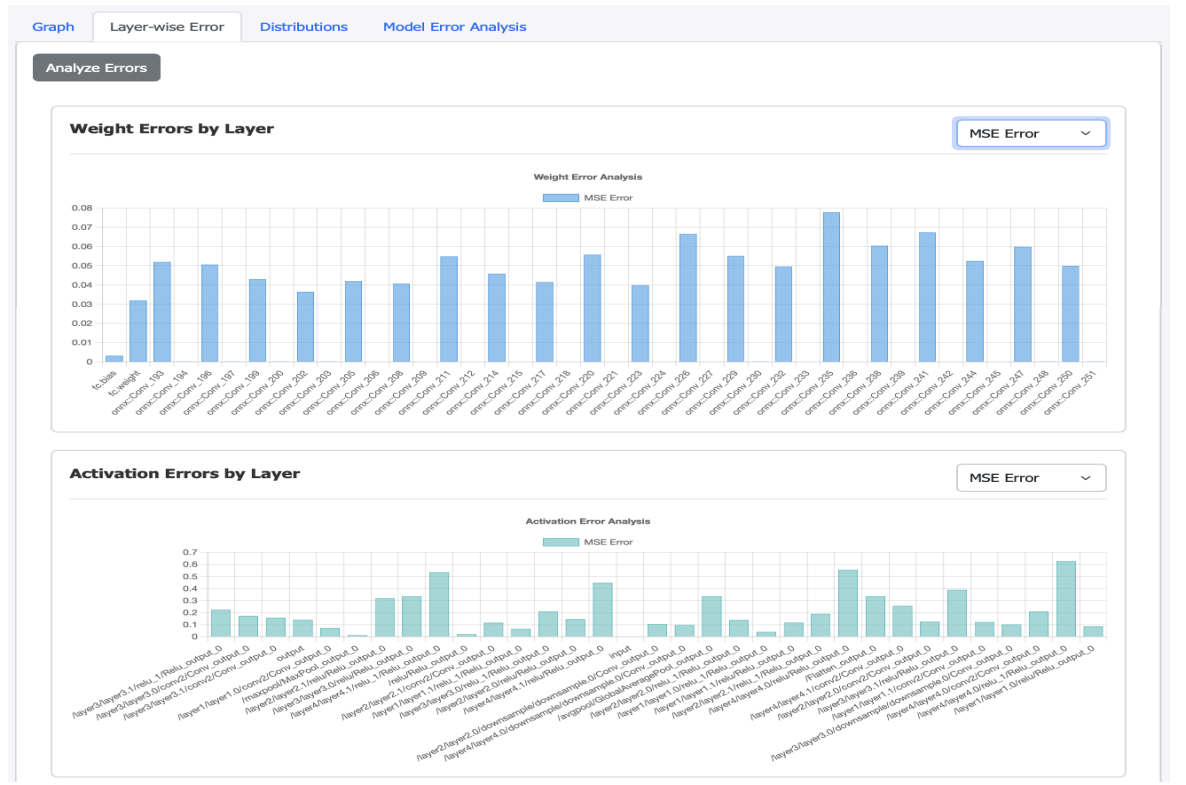

Fig. 3. Layer-wise sensitivity analysis visualization for weights and activations

### E. *Distribution Analysis of Weights and Activations*

To support interpretability of quantization behavior, the tool generates detailed histogram-based analyses for both weight tensors and activations generated during runtime. Weight distributions are extracted directly from FP32 model parameters and compared to their quantized counterparts to highlight dynamic range compression, clipping, and binning effects. Activation distributions are collected during calibration, reflecting real data flow patterns through the network as shown in Fig 4 for ResNet model.

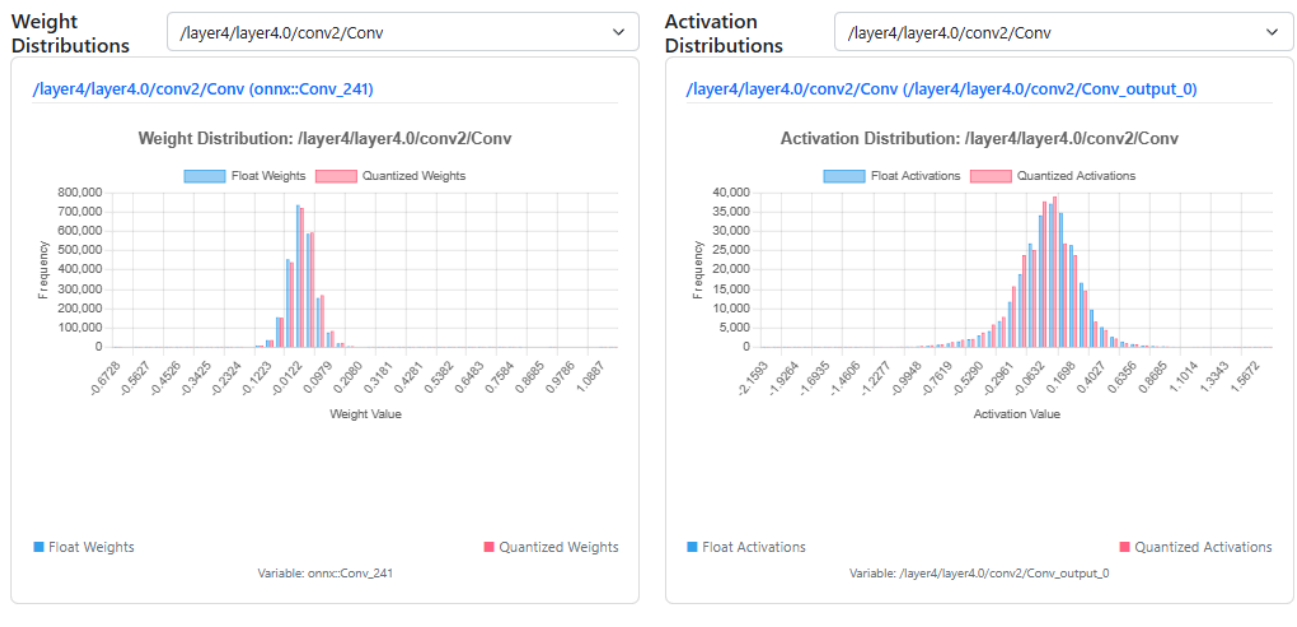

Fig. 4. Weight and Activation distributions visualization for float and quantized values

The tool visualizes these distributions using overlaid histograms that allow users to directly compare the floating-point and quantized representations for each layer. These visualizations make it possible to identify outlier sensitivity by highlighting extreme values that may be clipped during quantization, observe whether weights or activations are concentrated within narrow dynamic ranges, examine the behavioral differences introduced by asymmetric versus symmetric quantization schemes, and analyze layer-specific distribution skew that may influence quantization performance. Together, these insights provide a clear understanding of how numerical ranges and distribution shapes contribute to quantization-induced degradation. These insights help explain why certain layers exhibit high sensitivity and support targeted refinement.

Overall the tool provides actionable insights into the effects of quantization of model operations and its impact on the overall accuracy of the model. This is an essential step toward explainability and provides insights into model architecture design for quantization friendly and efficient deployment of AI models at the edge.

## IV. Results

Experiments were conducted across several architectures, provided in the torchvision collection of convolutional networks i.e., MobileNet, ResNet, EfficientNet, SqueezeNet and ShuffleNet, with emphasis on lightweight edge-oriented architectures. A standard calibration dataset (e.g., ImageNet subset) was used for analysis.

Analysis of the results highlights how the tool's integrated visual, statistical, and layer-wise diagnostic features directly improve the quantized accuracy of the models. As shown in Table 1, we see a significant improvement in quantized accuracy, by leveraging the insights from the quantization analysis tool. For models like ResNet, we see the standard PTQ quantization methods perform well as these models are well studied in the literature for quantization. The quantization becomes more challenging for smaller and more optimized models resulting in degradation in accuracy as seen for EfficientNet and SqueezeNet models. We identify the few (<10) sensitive layers in the model, which constitute less ~5% of the model operations to see a significant improvement in the quantized accuracy. Showing the tools is effective at identifying the sensitive layers. In the EfficientNet model we observe that excluding "Mul" operations improves the accuracy significantly, and the last "Gemm" operations are among the highly sensitive layers for quantization error. For SqueezeNet and ShuffleNet, layers at the early input stage are more sensitive. These insights provide valuable information to the model developers, also working across different architectures, which might need different solutions based on the weight/activation distributions and architecture differences, instead of one solution fits all approaches.

TABLE I. Quantized Accuracy analysis of different model architectures

| Model | Baseline Accuracy | PTQ | Tool Guided PTQ |
|---|---|---|---|
| ResNet-18 | 69.5 % | 69.0 % | 69.0 % |
| ResNet-50 | 80.0 % | 77.2 % | **77.4** % |
| MobileNet-V2 | 70.7 % | 67.3 % | **68.7** % |
| MobileNet-V3-Large | 74.1 % | 69.2 % | **71.3** % |
| EfficientNet-B0 | 77.1 % | 37.4 % | **54.9 %** |
| EfficientNet-B1 | 77.4 % | 73.5 % | **73.9** % |
| SqueezeNet 1.0 | 58.6 % | 57.6 % | **58.3 %** |
| SqueezeNet 1.1 | 59.5 % | 58.9 % | 58.8 % |
| ShuffleNet V2 0.5 | 60.4 % | 43.4 % | **57.0 %** |
| ShuffleNet V2 1.0 | 68.6 % | 61.9 % | **65.6 %** |

Across models, several consistent patterns emerged. Early convolutional layers displayed heightened sensitivity to quantization for SqueezeNet and ShuffleNet models, likely because they encode foundational spatial features that propagate through the entire network, making even small perturbations consequential. Depthwise convolutional layers showed more variable behavior in MobileNet models, with their sensitivity driven heavily by the shape and spread of their activation distributions. In contrast, fully connected layers near the output showed higher sensitivity for EfficientNet architecture, suggesting skipping last stage fully connected operations may improve accuracy of quantized models. Visual examination of weight and activation histograms further clarified these trends: layers that required higher precision tended to exhibit skewed or heavy-tailed distributions, while layers with compact, near-Gaussian distributions proved more stable under quantization. Activation distributions also help understand why these operations are highly sensitive to quantization, as large outlier distributions may lose representational range and are more prone to quantization-induced distortion. The layer sensitivity analysis consistently shows that weight errors are significantly lower compared to activation errors, emphasizing the unpredictable nature of activation and the importance of a good calibration dataset and calibration method of effective quantization of the model.

Together, these findings explain why uniform quantization strategies often underperform compared to mixed-precision approaches that assign bit-widths based on layer-specific characteristics. Nevertheless, the analysis also reveals opportunities for improvement, including expanding support for more advanced clipping strategies, automating mixed-precision search, and incorporating hardware-specific performance models to better align quantization choices with real-world deployment constraints.

In addition, we also perform a node sensitivity analysis which quantizes nodes in the model in topological order, showing how the quantized accuracy and size of the model varies as quantized operations. Fig 6, we plot the normalized model accuracy and model size, when compared with the float model on a sample set of 100 ImageNet test images. For ResNet the model size decreases significantly without much impact on accuracy as seen in Fig 6 (a).

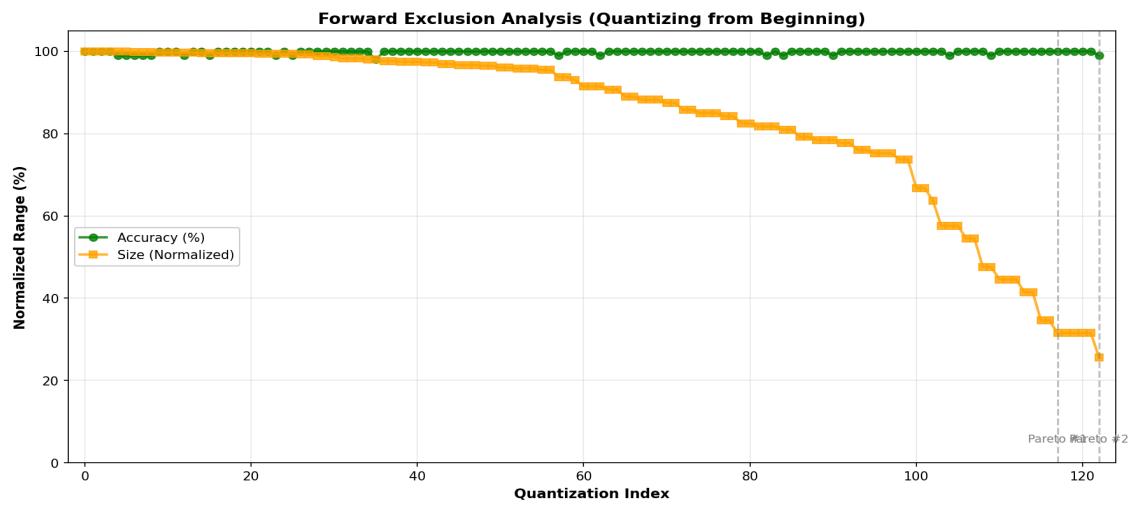

Fig 6. (a) Layer sensitivity for quantization in ResNet50 model

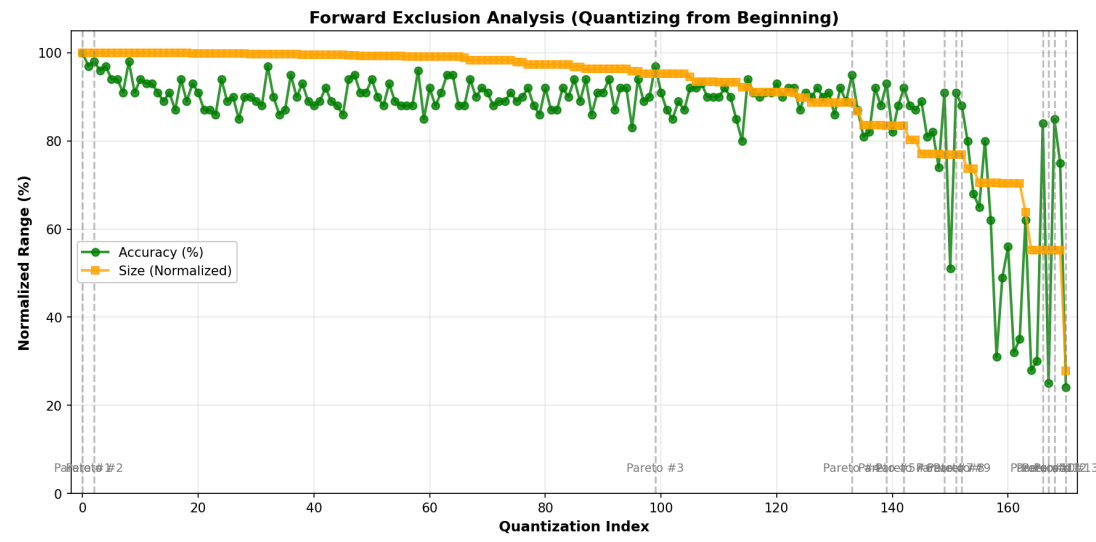

Fig 6. (b) Layer sensitivity for quantization in MobileNet V2

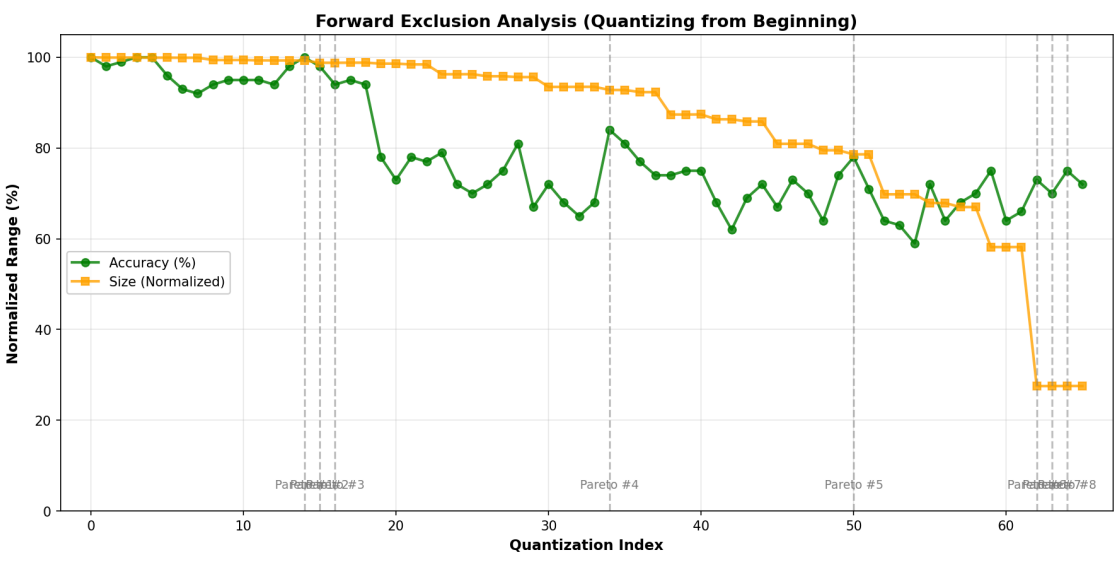

Fig 6. © Layer sensitivity for quantization in ShuffleNet 1.1

For MobileNet V2 we see a small drop in accuracy when quantizing the initial layers and a sudden dip in accuracy when quantizing the last few layers shown in Fig 6(b). Reinforcing the observation that the last few layers and initial layers can be critical for the quantized model accuracy. For ShuffleNet 1.1 we see a significant drop in the middle and doesn't drop significantly in the later stages of the model. This shows different characteristics of the model architecture and their weight distributions across different models.

## Conclusion

This paper introduced the Quantization Analysis Tool, a system for evaluating and optimizing ONNX-based quantized models. Through selective quantization experiments, the tool identifies sensitive layers, helping preserve accuracy while reducing model size. By combining layer-wise sensitivity analysis with detailed statistical visualizations, it offers actionable insights into quantization behavior. Experimental results show that tool-guided precision strategies outperform uniform post-training quantization, especially for complex models, providing a structured, data-driven approach to efficient AI deployment on edge devices.

Future work includes extending the tool to support advanced architectures such as Transformers and LLMs, adding new metric comparators for improved analysis, and evaluating quantized models on AI accelerators with limited INT8/INT16 or mixed-precision support. Given ONNX's wide adoption, these enhancements will further advance practical DNN quantization and help researchers and practitioners build efficient, robust AI systems for real-world deployment.

## Availability

The tool used for Analysis is available at: https://github.com/dwithchenna/onnx-analyer. For further information regarding the tool, please contact the authors.